\documentclass{article}
\usepackage[utf8]{inputenc}
\usepackage{comment}
\title{Neuro-evolutionary Frameworks for Generalized Learning Agents}
\author{Thommen George Karimpanal }
\date{February 2020}

\begin{document}

\maketitle
\begin{abstract}
    The recent successes of deep learning and deep reinforcement learning have firmly established their statuses as state-of-the-art artificial learning techniques. However, longstanding drawbacks of these approaches, such as their poor sample efficiencies and limited generalization capabilities point to a need for re-thinking the way such systems are designed and deployed. In this paper, we emphasize how the use of these learning systems, in conjunction with a specific variation of evolutionary algorithms could lead to the emergence of unique characteristics such as the automated acquisition of a variety of desirable behaviors and useful sets of behavior priors. This could pave the way for learning to occur in a generalized and continual manner, with minimal interactions with the environment. We discuss the anticipated improvements from such neuro-evolutionary frameworks, along with the associated challenges, as well as its potential for application to a number of research areas.
\end{abstract}

\section{Introduction}

The ultimate aim of artificial intelligence research is to develop agents with truly intelligent behaviors, akin to those found in humans and animals. To this end, a number of tools and techniques have been developed. In recent years, two approaches in particular - deep learning (DL) and reinforcement learning (RL), seem to have made considerable progress towards this goal.  
Both these fields have been widely studied, with numerous successful examples  \cite{kohl:icra04,Ng04invertedautonomous,Tesauro:1995:TDL:203330.203343,le2013building,taigman2014deepface} reported, particularly in recent years. However, even with the unprecedented success of recent approaches such as deep RL \cite{mnih2015human,mnih2013playing,silver2016mastering}, poor sample efficiency and limited generalization remain major concerns to be addressed, keeping in view the ultimate goal of developing general purpose agents. 

The poor generalization capability of DL is exposed by its liability to deception when presented with adversarial examples \cite{nguyen2015deep,szegedy2013intriguing}. Recent work \cite{su2019one}, showed that it was possible to hurt the performance of DL-based image recognition systems by carefully altering just a single pixel. Although DL is an extremely data-hungry approach, the issue of sample efficiency is perhaps more significant for RL systems, in which the time and energy costs associated with obtaining the samples via interactions with the environment can be prohibitively high. This is evident from the sparse number of cases where RL is directly applied to embodied systems. Traditional RL approaches require the learning to be carried out from scratch (tabula-rasa), and typically, in very large state-action spaces. The cost of exploring and learning in such spaces typically renders RL infeasible in many practical circumstances. Even in simulated environments such as Atari games \cite{bellemare2013arcade}, several thousands of training episodes and millions of environment interactions are typically needed in order to obtain acceptable agent behaviors in just a single game. Moreover, the resulting policy suffers from poor transfer properties, and is often limited to the particular task/game it was trained to solve. These issues pose major impediments to the practical utility of deep RL, particularly in embodied applications such as robotics. Such physical systems typically not have the luxury of enduring the consequences of such a large number of agent-environment interactions, especially when highly sub-optimal actions are occasionally chosen during the learning process. 

These drawbacks of DL and RL stand in stark contrast to how learning actually takes place in animals, who seem to acquire a diverse range of skills in a much more sample-efficient and generalized manner, seemingly, without any explicitly defined objective function. In addition, unlike in DL and RL, they do not learn tabula-rasa. They are born with several simple or elaborate priors, ingrained into their neural systems through millions of years of evolution and natural selection. 
Innate behaviors such as the suckling and grasping reflexes in human babies are examples of such priors.
Apart from innate behaviors, intrinsic mappings which help drive individuals towards survival, are also inherited as priors. For example, thirst, hunger and pain are mapped to negative states of being, which in turn drive individuals to take appropriate actions to escape these states. Other states such as warmth and satiety generally correspond to positive states of being, which individuals actively seek to attain. These intrinsic priors and mappings have a general applicability, in the sense that they are useful to the individual, irrespective of the tasks they go on to learn during their lifetime. Recent studies in humans \cite{dubey2018investigating} have shown that by directing exploration towards particular environmental states, such priors may also facilitate sample efficient learning. This naturally leads us to the questions:
\begin{itemize}
    \item What is the underlying mechanism that allows for the emergence of such diverse and generalized learning without the need for an explicit objective function?
    \item Can this process, along with prior acquisition mechanisms be mimicked in artificial agents to achieve sample-efficient learning?
\end{itemize}
In this paper, we qualitatively analyse these mechanisms and examine their potential applicability to artificial learning systems, and discuss how they may help achieve more generalized and sample-efficient learning, two of the necessary ingredients for artificial general intelligence \cite{goertzel2007artificial,bostrom2017superintelligence}. We also discuss the relevance of these mechanisms to a range of important research areas in artificial learning, along with some of the associated challenges.



\section{Acquisition of Generalized Skills via Relaxed Selection}
\label{relaxed_selection}

Generalization has been a longstanding problem in both DL and RL, both of which tend to produce solutions whose performance drastically deteriorates when evaluated on new tasks/environments.
The quest for generalization has driven entire fields of research such as feature engineering \cite{levine1969feature}, multi-task learning \cite{zhang2017survey,yang2017multi}, transfer learning \cite{pan2009survey,taylor_transfer_2009}, etc., all of which are concerned with the decoupling of task-specific characteristics from the skills that are generally useful to learn. A majority of these approaches essentially aim to extract and utilize domain-specific and task-independent knowledge that would likely be useful for a range of tasks within the domain. 
Although considerable progress has been made towards generalizing to task distributions \cite{du2019task,schaul2015universal,finn2017model}, the choice of these distributions is often arbitrary, and poorly justified. In addition, the domain knowledge that results from these approaches directly or indirectly relies on the optimization of a predefined objective function, the choice of which is again, usually poorly justified. 

Even the field of Evolutionary Algorithms (EA) \cite{holland1992genetic}, which was originally motivated by the problem of mimicking the emergence of generalized intelligent agent behavior, has failed in this regard. The idea behind EA is to initialize a population of `random' agents, and to subsequently maintain the population over a number of generations (iterations) through a fitness-proportional selection process. Following the selection process, the subsequent generation is populated by copying over a proportion of individuals directly as they are (exploitation), and by subjecting the remaining individuals to crossover and mutation operations (exploration). By selectively and repeatedly propagating fitter individuals to subsequent generations, the average performance of the population progressively improves, until finally, the desired agent behavior is achieved.
However, the problem with this approach is that the agent behaviors are optimized as per a specific and manually defined fitness function. This inevitably leads to a loss in the generality of the solutions, much like in RL and DL. In addition, manual specification of the fitness function can be counterintuitive, often leading to unexpected and undesired behaviors. 

In contrast to these approaches, natural agents (humans and animals) seem to be able to acquire a multitude of skills/behaviors (be it through learning or evolution), without an explicit objective function. 
Moreover, many of these skills seem to be general in nature, and applicable to a multitude of tasks. The apparent universal utility of these skills may seem to point to a violation of no free lunch theorems for optimization problems \cite{wolpert1997no}. However, this is not the case. As general as they may seem, the skills acquired by natural agents are applicable only to a subset of all possible problems. For example, certain fundamental skills acquired by humans, although beneficial for understanding language, and for complex motor skills such as walking, have no utility if the task is to communicate with other species, or to be able to jump over a tall building. This is because these are seldom encountered tasks in our natural environment, and the corresponding skills to solve them are generally not useful to acquire, in the sense that our survival and ability to reproduce/self-replicate would generally not be affected by the absence of these skills. Hence, the applicability of the skills acquired by natural agents seem to be limited to a subset of problems that are useful, in the sense as described above. Apart from usefulness, the environment also imposes feasibility constraints. For example, although it may be useful to control the motion of objects with just our thoughts, it is not possible to acquire such skills, as it violates the fundamental laws of physics, which are strictly imposed by the environment we inhabit. 

As it follows from the above arguments, from an evolutionary point of view, the only skills that seem to matter are those that are both feasible to acquire, and are directly or indirectly useful for the successful propagation of genetic information. This feasible usefulness is completely determined by the environment under consideration, and is a primary factor that ultimately guides the acquisition of skills. For example, for a given environment, only certain sets of traits may be feasibly useful. Individuals possessing these traits would be able to survive to be able to reproduce, in turn, making it likely for their very similar offspring to do the same. Unlike in traditional EA, where the selection is fitness proportional (as per a specifically defined fitness function), the selection criterion in natural evolutionary systems is decided by the environment under consideration, and is more relaxed, in the sense that it is merely dependent on whether the set of traits possessed by an agent allows it to be able to reproduce within its lifetime. Even simpler individuals which may correspond to `low fitness values' in the given environment, are allowed to self-replicate and propagate to the next generation, as long as they can survive for long enough to do so. As a natural consequence of this, the initial generations would be comprised of simpler individuals that are capable of self-replication, and as the generations proceed, more and more complex individuals would emerge, owing to mutations that occur during self-replication. This process would continue, producing a diverse set of agents (with a diverse range of skills and behaviors), all of which are adequately equipped for self-replication. 
The relaxed selection criterion imposed by the environment serves to implicitly encode a more generalized fitness function, which would eventually result in the generation of diverse sets of agent behaviors.

Despite the absolute autonomy made possible via relaxed selection, the underlying evolutionary mechanism is inherently slow, which might render the process infeasible for developing artificial agents. However, when deployed in combination with learning mechanisms, some of these drawbacks may be addressed by virtue of specific neuro-evolutionary mechanisms. 
A more detailed discussion of the nature of such neuro-evolutionary processes is presented in Section \ref{baldwinian}.

\section{From Learning to Inheritance - The Baldwin Effect}
\label{baldwinian}
Evolutionary mechanisms, in some sense, correspond to a form of inter-life transfer of knowledge, whereas learning is typically concerned with behaviors attainable within an individual's lifetime. However, in nature, the two mechanisms interact to produce interesting effects, particularly, the eventual acquisition of certain learned behaviors as evolutionary priors. The mechanism responsible for this phenomenon is called the Baldwin effect \cite{baldwin1896new}, and it provides an elegant explanation for how behaviors learned within a lifetime can eventually become instinctual over generations. 

The Baldwin effect can be be understood with a typical example of a predator-prey scenario. Suppose certain individuals in the prey population learn a particular predator-avoidance strategy, on average, more of these individuals would go on to survive and populate subsequent generations. As the generations proceed, individuals that learn this strategy earlier in their lives would survive for longer periods, thus making them more likely to survive to be able to reproduce more successfully. Due to natural selection, individuals possessing learning mechanisms with shorter and shorter learning times would be selected, until eventually, the learning time is so short that the behavior effectively becomes instinctual. 

This phenomenon has also been studied in artificial learning systems \cite{hinton1987learning,ackley1991interactions,fernando2018meta}. Hinton and Nowlan \cite{hinton1987learning} showed how the processes of evolution and learning interact to accelerate the acquisition of desired behaviors via the Baldwin effect. It was shown to be particularly useful in cases where the fitness function was highly non-smooth, which is the case with many tasks in nature. However, previous studies were based on traditional EA, which probably restricted the generalization of the solutions. In principle, Baldwinian effects would also arise when learning is combined with relaxed selection-driven evolutionary mechanisms (Section \ref{relaxed_selection}). Such a system could potentially result in diverse sets of generalized behaviors eventually being inherited as priors, which could potentially equip agents to learn a range of useful behaviors with minimal interactions with the environment.

To demonstrate this phenomenon, consider a harsh (scarce food, presence of predators etc.,) simulated environment in which artificial agents are equipped with RL-like learning mechanisms. For simplicity, let us assume that self-replication, food consumption and predator avoidance are some of the possible agent behaviors that can be learned in such an environment. A natural consequence of applying the relaxed selection criterion would be the emergence of an initial population of learning agents with reward structures that have an affinity for self-replication. This is because agents who are born with reward structures that are averse to self-replication would automatically be selected against, and would eventually die out. Within this initial population, there could be considerable variation in the agents' individual behaviors, in terms of their relative affinities for the learnable behaviors mentioned above. For example, individuals with greater affinity for self-replication would naturally cause similar agents to become proportionately more abundant in the population. In subsequent generations, affinities that enable agents to survive for longer periods may be discovered, and would propagate through the population. This is because agents that learn to survive for longer periods, say, by learning to forage and/or avoid predators, would probably produce a greater number of offspring, owing to the longer period in which it could potentially self-replicate. In some sense, learned behaviors such as foraging and predator avoidance simply serve to make the agents more efficient at self-replication. In reality, the number of offspring produced is not the only measure of replication efficiency. How reliably these offspring reproduce would also matter. The process thus incentivises frequent and reliable self-replication.

As the process continues, one could imagine the automatic inheritance of complex learned behaviors that serve to directly or indirectly improve the replication efficiency. For example, as the generations proceed, it may lead to the emergence of agents that are equipped with the necessary priors to quickly and easily learn to build tools for more efficient foraging, construct enclosures to protect themselves against predators, communicate with each other to exchange survival strategies, etc., It is worth noting that the emergence of these behaviors is rooted in the Baldwin effect, coupled with the implicit fitness function specified by the relaxed selection criterion.

\section{Research Potential}
The unique effects of relaxed selection, operating in conjunction with the Baldwin effect could uncover novel frameworks for the design of artificial agents with generalized behaviors. 
Such frameworks could possibly address and resolve a number of issues currently plaguing the field of artificial intelligence. We discuss some of these issues, with the hope that the core idea of enabling the autonomous acquisition of behaviors and priors through the interaction of learning and evolution 
will inspire future attempts at developing more general purpose agents:
\subsection{Freedom From Objective Functions}
One of the principal benefits of using the described approach would be that the design of agents would no longer be tied down to arbitrary and pre-defined fitness or objective functions. As relaxed selection allows the self-replication of any agent that can survive for long enough to do so, the focus would need to shift from designing the perfect fitness function to simulating the environment as closely as possible. 
Results from the field of model-based learning \cite{ha2018world,polydoros2017survey,kaiser2019model}, which is dedicated to the cause of building more accurate environment models, could provide a useful starting point for this undertaking. However, the described neuro-evolutionary frameworks would entail more high-resolution, detailed models that can account for various aspects of real world interactions. Although this is likely to be a highly computationally intensive task, with the proliferation of cheaper and more powerful computing resources, the idea may not be too far-fetched. Even with existing computational capabilities, novel concepts along the lines of observational dropout~\cite{freeman2019learning} could enable sample-efficient ways of obtaining models, with emphasis on their usefulness, rather than mere accuracy.  
An alternative would be to deploy these mechanisms on physical systems such as field robots in the real world. This would circumvent the need for developing computationally intensive, and possibly inaccurate simulated environments. However, it would translate to drastically increased training times, as well as other issues such as wear, high cost and other limitations, as is the case in Evolutionary robotics~\cite{nolfi2016evolutionary}.

\subsection{Intrinsic and Innate Behaviors}
One of the primary limitations of DL and RL algorithms is that the learning generally occurs from scratch, without appropriate initial behaviors that aid learning in the long run. Topics such as intrinsic motivation \cite{singh_intrinsically_2010,chentanez_intrinsically_2004} and transfer learning \cite{taylor_transfer_2009} which have garnered interest in the learning community, attempt to partially address this issue. However, they are limited by the requirement of arbitrary task distributions and/or directly or indirectly specified, but often poorly justified objective functions. The unique properties of relaxed selection, and the corresponding inheritance of prior behaviors via the Baldwin effect could lay the foundation for novel approaches to acquire sets of generalized priors and mappings, which could translate to several innate behaviors that are beneficial for self-replication. It would be interesting to study this effect in artificial systems, and to analyze the transformation of learned behaviors into intrinsic ones. It also opens up new avenues for the study of other related topics such as the nature of behaviors which are likely to become innate, the conditions under which innate/instinctual behaviors arise, the type of tasks which may benefit most from priors and mappings, etc., 



\subsection{Automation of Task Hierarchies}
An artificial agent deployed in the real world may be required to learn and manage a number of tasks. In many cases, the tasks may be interdependent \cite{kowsari2017hdltex,kulkarni2016hierarchical}, such that specific tasks or sets of tasks may need to be completed before others can be undertaken. For example, the task of cooking can only be undertaken following the successful completion of the task of procuring food and other materials. This may in turn be dependent on other tasks such as farming or hunting. The idea is that there exists a tree of task hierarchies which the agent must be aware of in order to evaluate the feasibility of a given task. With relaxed selection, the hypothesis is that self-replication lies at the root of a hierarchical evolutionary reward structure, and any behavior (such as food gathering, learning to ward off threats, developing energy-efficient technologies etc.,) that aids this ultimate objective, is assigned partial credit, which is encoded into the reward structure for the corresponding learning mechanism. 

The study of this evolutionary credit-assignment problem \cite{whitacre2006credit} could be used for extracting task dependencies, which could in turn be used to equip agents with intent, enabling them to autonomously choose their own tasks, or decide which task or set of tasks should be undertaken next. This is an aspect that is generally ignored by current approaches, which generally do not allow room for automated task identification~\cite{KARIMPANAL201739} and selection. Although the idea of artificial intent could be very useful for learning agents, the permitted set of intentions must be strictly restricted, and perhaps even monitored with human oversight. Automated intent could also be coupled with existing approaches for safe learning \cite{amodei2016concrete}. Such an integration may inform the development of the type of safety algorithms that need to be used in conjunction with truly autonomous agents of the future.

In addition, the hypothesis of self-replication efficiency being at the root of a hierarchical reward structure is implicitly related to the idea of explainability~\cite{gunning2017explainable,chandrasekaran1989explaining}, as such a structure may help dictate why a certain solution was proposed or why a certain task was assigned a higher priority.

\subsection{Learning Efficient Representations}
During an agent's lifetime, it may encounter a number of tasks. Some of these may be fairly simple, while others may be more complex. A wide spectrum of task complexities motivates a corresponding variation in the representations~\cite{bengio2013representation} used for learning. As relaxed selection operates on the basis of usefulness for self-replication, it may naturally allow for the emergence of sets of representations that are likely to be useful during the agent's lifetime. Representations that directly or indirectly boost the replication efficiency would be favored, while others would be selected against. This could potentially lead to the discovery and utilization of novel representations that are equipped to perform tasks such as causal reasoning~\cite{pearl2019seven}. In addition, relaxed selection could perhaps also give rise to mechanisms to select appropriately complex representations commensurate with the significance or complexities of different tasks. Acquiring generalized representation systems in this manner may also allow for the seamless transfer of knowledge across tasks.

\section{Challenges}
Although the combination of learning algorithms with relaxed selection mechanisms seems to be a promising approach for the realization of various facets of general intelligence, there are a number of challenges that may need to be overcome. Existing learning mechanisms, for instance, may not be sufficiently flexible for the emergence of certain aspects of generalized behaviors, such as learning task hierarchies. For such behaviors, 
more dynamic mechanisms that can encode ways for learning networks to split, merge and communicate with other networks may be needed.

Unlike in traditional EA, where the agent population remains constant across generations, a natural consequence of relaxed selection is that the agent population would continue to grow exponentially. As a consequence, it may quickly become infeasible to evaluate all agents in the subsequent generation. It may thus be necessary to develop methods to evaluate only a representative subset of the population. Alternatively, periodic artificial extinction events \cite{karimpanal2018self} may need to be introduced in order to restrict the population size.

Apart from the computational aspects, allowing behaviors to emerge naturally by means of relaxed selection mechanisms has the disadvantage  that the process is completely autonomous, and thus, may not be controllable. Successful, but undesirable behaviors that emerge may have to be weeded out frequently, either with appropriate safety mechanisms or using human oversight.

Despite these challenges, the fact remains that generalized behaviors entail generalized objectives, which may be implicitly encoded via the framework discussed in this work. To this end, evolutionary mechanisms such as relaxed selection and the Baldwin effect are worth studying further, especially in an era marked by powerful and inexpensive computer hardware, and an ardent interest in equipping artificial agents with generalized capabilities. 


\bibliographystyle{plain}
\bibliography{example}

\end{document}